# A Fourier Domain Feature Approach for Human Activity Recognition & Fall Detection

*Asma Khatun, Sk. Golam Sarowar Hossain

Aliah University

*Abstract*—Commonly, the senses of vision and hearing decrease as the age increases of a human. The most affected organs are hearing and vision due to aging. Elder people consequence a variety of problems while living Activities of Daily Living (ADL) for the reason of age, sense, loneliness and cognitive changes. These cause the risk to ADL which leads to several falls. Getting real life fall data is a difficult process and are not available whereas simulated falls become ubiquitous to evaluate the proposed methodologies. From the literature review, it is investigated that most of the researchers used raw and energy features (time domain features) of the signal data as those are most discriminating. However, in real life situations fall signal may be noisy than the current simulated data. Hence the result using raw feature may dramatically changes when using in a real life scenario. This research is using frequency domain Fourier coefficient features to differentiate various human activities of daily life. The feature vector constructed using those Fast Fourier Transform are robust to noise and rotation invariant. Two different supervised classifiers kNN and SVM are used for evaluating the method. Two standard publicly available datasets are used for benchmark analysis. In this research, more discriminating results are obtained applying kNN classifier than the SVM classifier. Various standard measure including Standard Accuracy (SA), Macro Average Accuracy (MAA), Sensitivity (SE) and Specificity (SP) has been accounted. In all cases, the proposed method outperforms energy features whereas competitive results are shown with raw features. It is also noticed that the proposed method performs better than the recently risen deep learning approach in which data augmentation method were not used.

I. INTRODUCTION & BACKGROUND

According to the American Center for Disease Control and Prevention (CDC), falls are the leading cause of fatal and nonfatal injuries among adults aged ≥65 years (older adults) [1-4]. They found that, during 2014, approximately 27,000 older adults died because of falls; 2.8 million were treated in emergency departments for fall-related injuries, and approximately 800,000 of these patients were subsequently hospitalized. The CDC researchers also investigated that every year the rates of fall-related deaths in the US have increased by 30% [5]. If this rate continues to increase, seven fall-related deaths can be estimated every hour by 2030 [6] in the U. S. It has become very important to develop fall detection algorithm which can automatically monitor and detect fall. Although there are number of research exists to prevent fall detection by the medical assessment through vitamin D supplication, advising to regular exercise and regular risk assessment [3, 7, 8], there are still major chances to get fall at the adult stage and furthermore, they need to be quickly detect and treat to prevent serious injury to the fallen victim. Moreover, the risk of fall occurrence increases for visual impairment people. To overcome this major challenge, earlier used direct Personal emergency response systems (PERS) such as [10, 11] are the commercial solutions. In that system, the victimized person can use a press button system to contact an emergency center for help. However, in many situations, the PERS system is not much useful when the person was alone or lay down on the floor for long time or unconscious which may not be able to reach the button and they need help to get up. Even, a recent cohort study found that around 80% of older adults wearing a PERS did not use their alarm system to call for help after getting a fall despite the alarm system was installed nearby [11]. It may be due to the fact that the victim was alone or unconscious. The condition becomes more serious when the victim has cognitive impairment due to for example aging and sex. Additionally, women are more likely to less get up than male from fall. Moreover, the older adults who lie longer and fall alone was not able to ask for their help and who are victimized with serious injuries causes to admission to hospital and long term medication. Moreover in some situation with other complicated patients such as heart patients it may cause more serious trouble. Because of these aforementioned challenges occur in direct PERS system, passive monitoring systems have been investigated to detect falls more sophisticated and accurate manner. In the literature the passive monitoring system has major three classifications such as wearable sensor based, camera (vision) based and non-wearable based which are depend on sensor technology. Compared to the earlier (direct PERS) these above mentioned methods are newest in the literature. In wearable sensor based system data are taken by using an accelerometer, a gyroscope and others like magnetometer. Whereas in non-wearable based technology, vibration or infrared sensors are used. In vision based method data acquisition are performed by set of video camera monitored system. The benefit of wearable devices is of minimal computational cost compared to the others such as non-wearable based technology. In addition to that these types of devices are easy to set up installation but may cause undesirable situation for example device disconnection. Unlike wearable based method, for non-wearable sensor, it is less intrusive due to minimum interaction with patients but these are cost effective. For vision based system, the accuracy is high and lower intrusive but the computational cost is much higher and setting up installation is medium in complexity. A good and detail comparison summary of passive monitoring fall detection approaches are presented in the paper by Mubashir *et al* [12].

In summary, although wearable devices based fall detection approaches provide low detection accuracy these approaches are most popular than alternative approaches due to their cost effective and easy installation features. The pictorial representation of fall detection categories and their subcategories are represented in the Fig. 1 below.

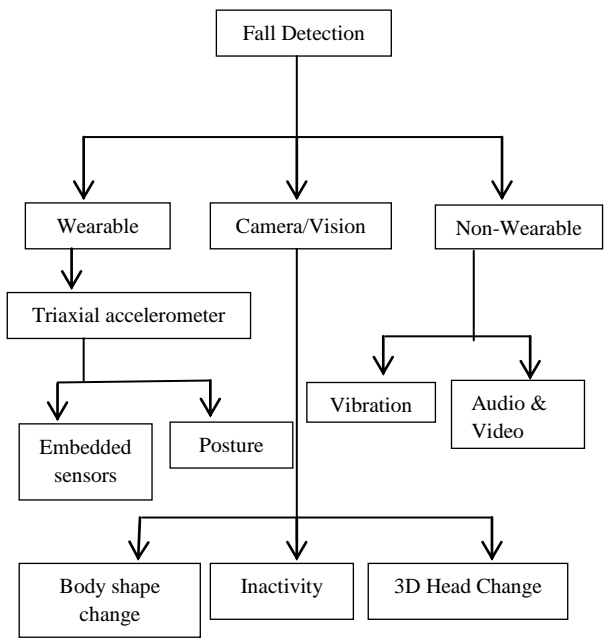

Fig. 1. Categories of passive moitoring fall detection method

In the literature, many works has been accomplished in fall detection. Most popular and up to date research will be discussed in this literature review section. Fall detection method can be broadly divided into threshold and machine learning based. Variety of features has been used to represent feature vector in the existing literature. Little researcher used deep learning based approaches. Table I presents some of those current approaches of the related work in brief.

This article presents a Fourier coefficients feature based fall detection approach based on smartphone accelerometer. The remainder of the paper is organized as follows. Section II provides the processing of data collection and the method design. Section III represents obtained experimental results and evaluations of the proposed method and comparison analysis. Finally, Section IV concludes the paper with future works.

TABLE I. LITERATURE REVIEW

| Approach | Method | Features |
|---|---|---|
| G. L. Santos et al. [13]$_{2019}$ | CNN | Deep CNN |
| Ramon et al. [14]$_{2018}$ | SVM, kNN, Naives Bayes & Decision tree | Mean, standard deviation of signal magnitude vector, mean rotation angle |
| Mezghani et al. [15]$_{2017}$ | SVM classifier | Max, min, range, mean, skewness, variances & orientation, |
| Micucci et al. [16]$_{2015}$ | One & two class kNN One & two class SVM | Raw acceleration data Magnitude Mean and standard deviation of the acceleratin values Eneregy and correlation of acceleration Local temporal patterns |
| Medrano et al. [17]$_{2014}$ | K-means & NN | Magnitude |
| Wang et al. [18]$_{2014}$ | Threshold based | Signal magnitude vector Hearth rate value Trunk angle |
| Rabah et al. [19]$_{2012}$ | Threshold based | Magnitude, orientation |
| Attal et al. [20]$_{2015}$ | kNN, SVM, GMM, RF, k-means & HMM | Raw features & selected from raw data |
| Gupta & Dallas [21]$_{2014}$ | Naïve Bayes & KNN | Energy, entropy, mean, variance, Max, mean trend, windowed mean & variance difference, X-Z energy. |

## II. METHODS & ANALYSIS

The proposed method consists of three main steps as data collection, feature vector representation and feature classification.

### A. Data Collection

Motivated by the researcher as in [16], we collected and divided dataset as described next. Smartphone accelerometer data are collected from two publicly dataset [17] and [22]. Those data are of heterogeneous devices and setup installation. In brief, Medrano et al. [17] experimented with 8 types of simulated falls and ADL of 10 volunteers of forward falls, backward falls, left-lateral and right-lateral falls, syncope, sitting on empty chair, falls using compensation strategies and falls with obstacle. On the other hand Anguito et al. [22] recorded 16 types of ADL of 30 subjects carrying a waist-mounted smartphone with embedded inertial sensors. Some examples of acceleration signals obtained during falls and ADL are provided in the Figure 2 below. These two above datasets are mixed and separated into three collection of data as follows [16].

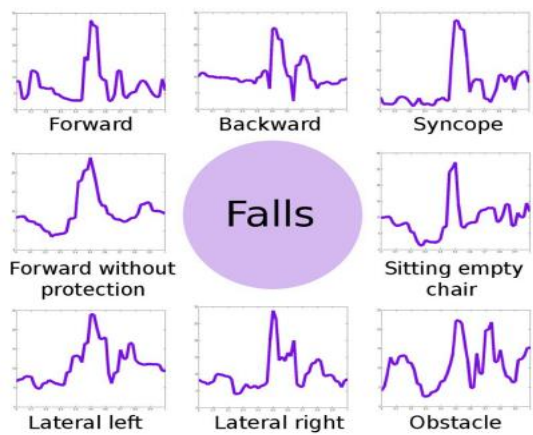

(a)

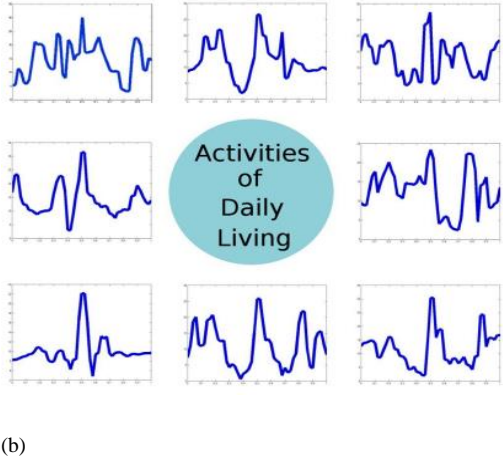

(b)

Fig. 2. Some examples of acceleration shapes obtained during (a) falls and (b) ADL, Images adapted from Medrano *et al.* [17].

- **Collection 1.** ADL: 7035 training and 781 test data. FALL: 453 training and 50 test data. Both ADL and FALL data have been collected from the dataset1.
- **Collection 2.** ADL: 7035 training and 781 test data. Half of ADL data have been randomly collected from the dataset1 and half from the dataset2; FALL: 453 training data and 50 test data. All the FALL data have been collected from the dataset1.
- **Collection 3.** ADL: 9270 training data and 1029 test data. All the ADL data have been taken from the dataset2; 453 FALL training data and 50 FALL test data. All the FALL data have been collected from the dataset1.

During the data processing, 90% data are used as a training data and 10% as testing data from the above data collection.

### B. Feature Vector Representation & Classification

In real life scenario fall are non-simulated and the existing time domain based feature such as raw, energy, magnitudes features might be error prone to noisy data. Hence this research proposed a new feature vector using Fourier coefficients in frequency domain. It is to be noted that frequency domain features are less noisy and robust. After that SVM and kNN (k nearest neighbor) classifier are used to classify the ADL and fall instances.

### III. EXPERIMENTAL EVALUATION & RESULT ANALYSIS

The experiment is conducted using both one class and two class classifier with 10 fold cross validation technique. The research is also tested with both 51 and 128 window sample size and observed that using 128 sampling window size the better results provide. There are mainly tree evaluation metrics including threshold metrics, ranking metrics and probability metrics used in the regard of machine learning classification approaches. Among the above first two metrics are suitable in class imbalance problem. Following evaluation metrics such as *Standard Accuracy, Macro Average Accuracy* [23], *Sensitivity* and *Specificity* are used to evaluate the proposed method:

1) Standard accuracy a global measurement evaluation method, which is a multi class ranking metrics and is described as follows: Given E the set of all the activities types, $a \in E$, $NP_a$ the number of times a occurs in the dataset, and $TP_a$ the number of times the activity a is recognized:

$$\text{Standard Accuracy} = SA = \frac{\sum_{a=1}^{|E|} TP_a}{\sum_{a=1}^{|E|} NP_a} \quad (1)$$

2) Macro average accuracy is also a multiclass thresholds metrics which is suitable for data imbalance and is defined as follows:

$$\text{Macro Average Accuracy} = MAA = \frac{1}{|E|} \sum_{a=1}^{|E|} A_{cc_a}$$

$$= \frac{1}{|E|} \sum_{a=1}^{|E|} \frac{TP_a}{NP_a} \quad (2)$$

MAA is the arithmetic average of the accuracy $A_{cc_a}$ of each activity. It allows each partial accuracy to contribute equally to the evaluation.

Other most popular evaluation metrics such as Sensitivity and Specificity are described as follows: SN and SP are one class threshold metrics. The one of the major advantage SN and SP are that they are not affected by data imbalance.

3) Sensitivity is obtained by $SE = \frac{True\ Positives}{Positives}$ (3)

4) Specificity is obtained by $SP = \frac{True\ Negatives}{Negatives}$ (4)

Below the results are discussed. Results of the 10 fold cross evaluations with both SVM and kNN are presented in the following Tables II, III, IV, V & VI. This research is mainly compared with two feature raw and energy as those two features are most common and discriminating which are described in the Table I. Table II presents comparison results using energy feature and the proposed Fourier coefficients feature vectors with Collection 1 datasets measuring SA and MAA. Table III provides the comparison result using energy and frequency feature based on Collection 2 dataset and Table IV presents the comparison results based on Collection 3 dataset using the same energy and frequency feature measuring SA and MAA respectively. Table V & VI shown the results using SE and SP measurement of Collection 1 & Collection 2 datasets respectively.

From the experiments it is investigated that kNN classifier performs better than SVM classifier hence all the results are detailed with kNN classifier approach.

TABLE II. COMPARION RESULTS USING ENERGY AND THE PROPOSED FOURIER COEFFICIENTS FEATURES IN COLLECTION 1 DATASET

| Two classes kNN energy feature | Two-classes kNN Fourier coefficients Features |
|---|---|
| SA: 95.31  MAA: 79.72 | SA: 93.14  MAA: 73.89 |
| SA: 92.78  MAA: 74.63 | SA: 94.46  MAA: 75.53 |
| SA: 94.22  MAA: 76.34 | SA: 94.71  MAA: 74.72 |
| SA: 94.46  MAA: 71.78 | SA: 93.26  MAA: 69.27 |
| SA: 93.86  MAA: 75.21 | SA: 93.98  MAA: 73.40 |
| SA: 93.14  MAA: 70.14 | SA: 94.58  MAA: 77.46 |
| SA: 93.98  MAA: 74.34 | SA: 94.83  MAA: 74.78 |
| SA: 93.14  MAA: 72.02 | SA: 94.58  MAA: 77.46 |
| SA: 95.31  MAA: 69.42 | SA: 94.34  MAA: 73.59 |
| SA: 93.86  MAA: 72.40 | SA: 94.34  MAA: 77.34 |
| mean_SA: **93.86** mean_MAA: **72.40** | mean_SA: **94.34** mean_MAA: **77.34** |

TABLE III. COMPARISON RESULTS IN COLLECTION 2 DATASET

| Two classes kNN energy feature | Two-classes kNN Fourier coefficients Features |
|---|---|
| SA: 98.05  MAA: 85.66 | SA: 99.44  MAA: 95.90 |
| SA: 98.33  MAA: 85.81 | SA: 99.44  MAA: 94.95 |
| SA: 97.59  MAA: 80.66 | SA: 99.35  MAA: 93.95 |
| SA: 97.59  MAA: 85.42 | SA: 99.35  MAA: 95.85 |
| SA: 97.78  MAA: 83.61 | SA: 98.80  MAA: 88.90 |
| SA: 97.50  MAA: 85.37 | SA: 99.63  MAA: 97.90 |
| SA: 98.33  MAA: 87.71 | SA: 99.54  MAA: 96.90 |
| SA: 97.68  MAA: 85.47 | SA: 99.44  MAA: 96.85 |
| SA: 98.33  MAA: 88.66 | SA: 99.54  MAA: 95.00 |
| SA: 97.96  MAA: 84.66 | SA: 99.54  MAA: 95.95 |
| mean_SA: **97.96** mean_MAA: **84.66** | mean_SA: **99.54**  mean_MAA: **95.95** |

TABLE IV. COMPARISON RESULTS IN COLLECTION 3 DATASET

| Two classes kNN energy feature | Two-classes kNN Fourier coefficients Features |
|---|---|
| SA: 93.38  MAA: 76.82 | SA: 94.95  MAA: 73.91 |
| SA: 95.31  MAA: 75.98 | SA: 94.83  MAA: 72.91 |
| SA: 94.10  MAA: 74.40 | SA: 95.67  MAA: 81.78 |
| SA: 93.50  MAA: 72.21 | SA: 95.31  MAA: 74.10 |
| SA: 93.98  MAA: 64.98 | SA: 94.83  MAA: 75.72 |
| SA: 96.15  MAA: 81.10 | SA: 93.98  MAA: 72.46 |
| SA: 94.46  MAA: 76.46 | SA: 94.10  MAA: 76.27 |
| SA: 94.58  MAA: 75.59 | SA: 96.63  MAA: 86.98 |
| SA: 95.19  MAA: 75.91 | SA: 96.27  MAA: 82.10 |
| SA: 94.34  MAA: 75.46 | SA: 96.03  MAA: 82.91 |
| mean_SA: **94.34** mean_MAA: **75.46** | mean_SA: **96.03** mean_MAA: **82.91** |

TABLE V. COMPARISON RESULTS IN COLLECTION 1 DATASET USING SE AND SP MEASURES

| One class kNN energy feature | One-class kNN Fourier coefficients Features |
|---|---|
| SE: 73.96  SP: 76.74 | SE: 65.41  SP: 81.11 |
| SE: 65.01  SP: 87.48 | SE: 67.40  SP: 80.12 |
| SE: 77.34  SP: 75.15 | SE: 69.38  SP: 77.93 |
| SE: 73.56  SP: 75.94 | SE: 72.37  SP: 77.34 |
| SE: 72.37  SP: 79.72 | SE: 70.38  SP: 78.13 |
| SE: 63.42  SP: 85.88 | SE: 68.99  SP: 78.53 |
| SE: 65.81  SP: 85.09 | SE: 68.99  SP: 79.92 |
| SE: 63.02  SP: 86.28 | SE: 65.01  SP: 81.31 |
| SE: 66.20  SP: 87.08 | SE: 66.20  SP: 78.53 |
| SE: 77.73  SP: 75.94 | SE: 68.19  SP: 81.71 |
| mean_SE: 77.73   mean_SP: **75.94** | mean_SE: 68.19   mean_SP: **81.71** |

TABLE VI. COMPARISON RESULTS IN COLLECTION 2 DATASET USING SE AND SP MEASURES

| One class kNN energy feature | One-class kNN Fourier coefficients Features |
|---|---|
| SE: 95.63  SP: 89.07 | SE: 98.01  SP: 96.02 |
| SE: 92.45  SP: 91.05 | SE: 96.82  SP: 97.61 |
| SE: 95.03  SP: 86.68 | SE: 97.81  SP: 97.22 |
| SE: 96.62  SP: 86.48 | SE: 97.22  SP: 96.22 |
| SE: 95.83  SP: 85.09 | SE: 95.63  SP: 98.41 |
| SE: 93.44  SP: 87.48 | SE: 98.61  SP: 96.42 |
| SE: 96.22  SP: 87.28 | SE: 96.62  SP: 98.41 |
| SE: 94.43  SP: 90.06 | SE: 98.21  SP: 96.02 |
| SE: 94.43  SP: 89.07 | SE: 98.41  SP: 96.02 |
| SE: 92.84  SP: 92.05 | SE: 96.82  SP: 98.21 |
| mean_SE: **92.84**   mean_SP: **92.05** | mean_SE: **96.82**   mean_SP: **98.21** |

Results indicate that the proposed method using Fourier coefficients outperforms the so called energy feature in all measurement. It is noteworthy to be mentioned that the proposed method gained maximum of *96.82* of Sensitivity and *98.21* Specificity which is also higher than the recently proposed deep learning approach without data augmentation method, proposed by Santos *et al.* [13]. They have achieved *83.33* and *87.50* in cases of SN and SP respectively by the method of using deep learning and without data augmentation. However they have received more promising results applying data augmentation technique on the deep learning approach and get 99.72 and 100 SN and SP respectively. It is to be noted that deep learning approach works well for large scale database (such as million). Hence the question arises on datasets integration method. Moreover, they used homogenous URFD (video) dataset which have limited number of data such as 30 falls and 40 ADLs. In their process, 5000 new samples are augmented. In addition to that deep learning based methods are computationally expensive than the conventional feature based method. On the other hand, feature based approach are more suitable while small dataset used. On the other hand our experiment also tested with raw feature and competitive results have gained. Some of the comparison results are shown in the Table VII and Table VIII.

TABLE VII. COMPARION RESULTS USING RAW AND THE PROPOSED FOURIER COEFFICIENTS FEATURES IN COLLECTION 1 DATASET

| Two classes kNN raw feature | Two-classes kNN Fourier coefficients Features |
|---|---|
| SA: 97.59  MAA: 86.55 | SA: 93.14  MAA: 73.89 |
| SA: 97.83  MAA: 82.94 | SA: 94.46  MAA: 75.53 |
| SA: 97.95  MAA: 83.94 | SA: 94.71  MAA: 74.72 |
| SA: 98.19  MAA: 85.94 | SA: 93.26  MAA: 69.27 |
| SA: 97.71  MAA: 81.94 | SA: 93.98  MAA: 73.40 |
| SA: 97.11  MAA: 76.94 | SA: 94.58  MAA: 77.46 |
| SA: 97.71  MAA: 82.87 | SA: 94.83  MAA: 74.78 |
| SA: 97.23  MAA: 79.81 | SA: 94.58  MAA: 77.46 |
| SA: 97.83  MAA: 82.94 | SA: 94.34  MAA: 73.59 |
| SA: 97.11  MAA: 79.74 | SA: 94.34  MAA: 77.34 |
| mean_SA: **97.11** mean_MAA: **79.74** | mean_SA: **94.34** mean_MAA: **77.34** |

TABLE VIII. COMPARISON RESULTS USING RAW AND THE PROPOSED FOURIER COEFFICIENTS FEATURES IN COLLECTION 2 DATASET

| Two classes kNN raw feature | Two-classes kNN Fourier coefficients Features |
|---|---|
| SA: 99.63  MAA: 96.00 | SA: 99.44  MAA: 95.90 |
| SA: 99.54  MAA: 95.00 | SA: 99.44  MAA: 94.95 |
| SA: 99.44  MAA: 94.00 | SA: 99.35  MAA: 93.95 |
| SA: 99.35  MAA: 93.00 | SA: 99.35  MAA: 95.85 |
| SA: 99.35  MAA: 93.00 | SA: 98.80  MAA: 88.90 |
| SA: 99.72  MAA: 97.00 | SA: 99.63  MAA: 97.90 |
| SA: 99.63  MAA: 96.00 | SA: 99.54  MAA: 96.90 |
| SA: 99.35  MAA: 93.00 | SA: 99.44  MAA: 96.85 |
| SA: 99.54  MAA: 95.00 | SA: 99.54  MAA: 95.00 |
| SA: 99.72  MAA: 97.00 | SA: 99.54  MAA: 95.95 |
| mean_SA: **99.72**  mean_MAA: **97.00** | mean_SA: **99.54**  mean_MAA: **95.95** |

## IV. CONCLUSION

The advantage of the proposed method has two fold one is discriminating and the other one is robust to real life noise. The proposed method based on Fourier coefficients is discriminating as it has the property of robustness to the level of detail representation. Moreover, Fourier transform based descriptors are invariants under affine transformations which are composed of several transformations such as translation, rotation, flipping and scale. Our approach of using frequency domain shows competitive performances with raw based features (time domain) and outperforms energy based feature (time domain) vectors using the standard measurement methods such as SA, MAA, SE and SP. Moreover our method outperforms in the case of SE and SP measure of the recently proposed deep learning method in which data augmentation method was not used in their method. The proposed method is easy to implement and shows promising results. As a future work we would like to experiments with more evaluation metrics as well as using non-simulated data.